\documentclass{ijicic}
\usepackage{graphicx} 
\usepackage[margin=3cm]{geometry}
\usepackage[round, sort]{natbib}
\usepackage{hyperref}
\usepackage{abstract}
\usepackage{mathptmx} 
\usepackage{color}
 
\usepackage{courier}        
\usepackage{type1cm}        
\usepackage{graphicx,epsfig,color}
\usepackage{multicol}        

\usepackage{array,multirow}
 \usepackage{algorithm2e}
\usepackage{enumitem}
\usepackage{wrapfig}

\usepackage{url}
\usepackage{amsmath}
\makeatletter
\let\@noitemerr\relax
\makeatother

\title{{Accumulator Bet Selection through Stochastic Diffusion Search}}

\date{}
\begin{document}
\maketitle

\centerline{\scshape  Nassim Dehouche}
 \medskip
{\footnotesize
\centerline{Business Administration Division}
\centerline{Mahidol University International College}
\centerline{999 Phutthamonthon 4 Road, Salaya, Nakhonpathom 73170, Thailand}
\centerline{nassim.deh@mahidol.ac.th} }
\medskip

\centerline{Received May 2018; revised September 2018}


\medskip

\begin{abstract}
An accumulator is a bet that presents a rather unique payout structure, in that it combines multiple bets into a wager that can generate a total payout given by the multiplication of the individual odds of its parts. These potentially important returns come however at an increased risk of a loss. {Indeed, the presence of a single incorrect bet in this selection would make the whole accumulator lose.}
{The complexity of selecting a set of matches to place an accumulator bet on, as well as the number of opportunities to identify winning combinations have both dramatically increased with the easier access to online and offline bookmakers that bettors have nowadays}. We address this relatively under-studied combinatorial aspect of sports betting, and propose a binary optimization model for the problem of selecting the most promising combinations of matches, in terms of their total potential payout and probability of a win, to form an accumulator bet. The results of an ongoing computational experiment, in which our model is applied to real data pertaining to the four main football leagues in the world over a complete season, are presented and compared to those of single bet selection methods.

\end{abstract}


\section{introduction}
{The global sports betting market is worth an estimated 
\$700 billion annually \cite{MARKET}, and association football (also known as soccer or simply football), being the world's most popular spectator sport, constitutes around $70\%$ of this ever-growing market \cite{1}. The last decade has thus seen the emergence of numerous online and offline bookmakers, offering bettors the possibility to place wagers on the results of football matches {in more than a hundred different leagues, worldwide.}\\ 

The sports betting industry offers a unique and very popular betting product known as an \textit{accumulator bet}. In contrast {with} a single bet, which consists in betting on a single event for a payout equal to the stake (i.e. the sum wagered) multiplied by the odds set by the bookmaker for that event, an accumulator bet combines more than one (and generally less than seven) events into a single wager that pays out only when all individual events are correctly predicted. The payout for a correct accumulator bet is the stake multiplied by the product of the odds of all its constituting wagers. However, {if} one of these wagers is incorrect, the entire accumulator bet {would} lose. Thus, this product offers both significantly higher potential payouts and higher risks than single bets, and the large pool of online bookmakers, leagues and, matches {that bettors can access nowadays has increased} both the complexity of selecting a set of matches to place an accumulator bet on, and the number of opportunities to identify winning combinations.\\ 
With the rise of sports analytics, a wide variety of statistical models for predicting the outcomes of football matches have been proposed, a good review of which can be found in \cite{13}. Since, the classic work of \cite{MAHER} that forecasts match outcomes by modeling both {teams'} scores with a Poisson distribution, more recent models for the prediction of sports match outcomes mostly rely on Bayesian inference \cite{1, 2, 3, 4, 9} and prediction markets \cite{6, 7, 8, 10, 11}. Drawing on data from 678 to 837 of the German \textit{Bundesliga} clubs, \cite{11} showed that human tipsters are outperformed by both prediction markets and betting odds, in terms of forecasting accuracy. Rule-based reasoning was used in \cite{12}, in which historical confrontations of the two opponents are modeled using fuzzy logic, and a combination of genetic and neural optimization techniques is used to fine-tune the model. Bayesian inference was combined with rule-based reasoning in \cite{3} to predict individual football match results. This combination relies on the intuition that although sports results are highly probabilistic in nature, team strategies have a deterministic aspect that can be modeled by crisp logic rules, which has the added benefit of being able to generate recommendations, even if historical data are scarce. The original work presented in \cite{5} focuses on a single outcome for football matches, {the two teams drawing}, and proposes a single bet selection strategy relying on the Fibonacci sequence. Despite its simplicity, the proposed approach proves profitable over a tournament.  {However, to the best of our knowledge, existing models only offer recommendation on single bets and neglect an important combinatorial dimension in sports betting that stems from selecting accumulator bets.} {Section \ref{5} proposes a binary mathematical programming model to address this combinatorial problem.}\\ 

After this introductory Section, we define some relevant football betting concepts and further distinguish between the two main betting products offered by bookmakers, namely single and accumulator bets, in Section \ref{2}. The accumulator bet selection problem is mathematically formulated in Section \ref{4}, and Section \ref{5} introduces the binary mathematical programming model that we propose for its treatment. The results of an ongoing computational experiment, in which our model is applied to real data gathered from the four main football leagues in the world (Spain's \textit{LaLiga}, England's \textit{Premier League}, Germany's \textit{Bundesliga} and Italy's \textit{Serie A}), over a complete season are presented in Section \ref{6}. Lastly, we draw conclusions regarding the results of this experiment and perspective of further development of the proposed approach, in Section \ref{7}.  }
\section{{Proposed Model}}
\subsection{{Preliminary definitions}}
\label{2}

{A \textit{single bet} consists in betting on a single outcome, whereas an accumulator is a bet that combines multiple single bets into a wager that only generates a payout when all parts win. This payout is equal to the product of the odds of all individual bets constituting the accumulator. Thus, the advantage of an accumulator bet is that returns are much higher than those  of a single bet. This however comes at the expense of an increased risk. Indeed, only a single selection need to lose for the entire accumulator bet to lose}.\\ 

{Let us consider for the sake of illustration $k$ independent single bets (i.e. no bets on conflicting outcomes, for the same match). Each individual bet $i=1, \dots, k$ is a Bernoulli trial with success probability $p_i$ and success payout $o_i$. An accumulator would be another Bernoulli trial, when dividing a wager over these $k$ single bets would be the sum of $k$ independent, but not independently distributed Bernoulli trials. We compare the expected returns and variance (as a measure of risk) of these two ways to {place the bets}, namely:
\begin{itemize}
\item Wagering $1$ monetary unit on an accumulator bet constituted of these $k$ single bets would offer an expected return of $E(A)=\prod\limits_{i=1}^{k} o_i \cdot p_i $, with a variance of $V(A)=(\prod\limits_{i=1}^{k} o_i^2 \cdot p_i) \cdot (1-\prod\limits_{i=1}^{k}p_i)$. 
\item Wagering $\frac{1}{k}$ monetary unit on each one of the $k$ single bets, would offer an expected return of $E(S)=\frac{1}{k}\sum\limits_{i=1}^{k} o_i \cdot p_i $, with a variance of $V(S)=\frac{1}{k^2}\sum\limits_{i=1}^{k} o_i^2\cdot p_i\cdot (1-p_i)$.
\end{itemize}
If we assume, for the sake of simplification, that all individual bets present the same odds and probabilities, i.e. $o_i=o$ and $p_i=p, \forall i=1, \dots, k$, then the above-described accumulator bet would present expected returns of $E(A)=(o\cdot p)^k$, with a variance of $V(A)=(o^2 \cdot p)^k\times(1-p^k)$
when the single bets approach would present expected returns of $E(S)=o\cdot p$, with a variance of $V(S)=(o^2 \cdot p)\times \frac{1-p}{k}$, in which we can note that $(1-p^k)>(1-p)>\frac{1-p}{k}$.}

{Further, assuming that all single bets are in favor of the bettor and present expected positive returns $o\cdot p>1$, it can be easily seen, in this simplified setting, that the expected returns of an accumulator bet $(o\cdot p)^k$, are exponentially larger than those of a pool of single bets $(o\cdot p)$, but their significantly higher variance of $(o^2 \cdot p)^k\times(1-p^k)$, relative to $(o^2 \cdot p)\times \frac{1-p}{k}$ for a pool of single bets, reflects their riskier nature.}

\label{4}
On any given match day, bettors have access to up to $100$ leagues and $50$ bookmakers, proposing different odds. Thus, there exists a pool of around $10$ matches/leagues $\times$ $100$ leagues $\times$ $50$ bookmakers $\times$ $3$ possible outcomes/match = $1,500,000$ possible single bets, for each match day, and the possible combinations of these bets to constitute an accumulator bet is astronomical. 

For a given match day, we consider a set $M=\{m_1, m_2, \dots, m_n\}$ of $n$ matches, potentially spanning several leagues, in which each match can have one of three possible outcomes: \textit{Home win} (H), \textit{Draw} (D), or \textit{Away win} (A), we denote $p_i(H)$, $p_i(D)$, and $p_i(A)$ the respective estimated probability of each one of these outcomes, for each match $m_i: i\in\{1, \dots, n\}$.\\

We additionally consider a set $B=\{b^1, \dots, b^m\}$ of bookmakers, each bookmaker $b^j: j\in\{1, \dots, m\}$ offering certain odds $o_{i}^{j}(H)$, $o_{i}^{j}(D)$, and $o_{i}^{j}(A)$ for a correct individual bet on a home win, a draw, and away win, respectively, in match $m_i: i\in\{1, \dots, n\}$.\\      

One seeks to determine an accumulator bet, that is to say three exclusive subsets of matches $M^*(H)=\{m^*_{h_1}, \dots, m^*_{h_k}\}$, $M^*(D)=\{m^*_{d_1}, \dots, m^*_{d_k}\}$,  and $M^*(A)=\{m^*_{a_1}, \dots, m^*_{a_k}\}$, representing the matches to bet on for a home win, a draw, and an away win, respectively, and a bookmaker $b^{j^*} \in B$, such as to maximize the two following functions:
\begin{itemize}
\item The probability of a winning accumulator bet, given by: $$\displaystyle \prod_{i=h_1}^{h_k} p_i(H) \times \displaystyle \prod_{i=d_1}^{d_k} p_i(D) \times \displaystyle \prod_{i=a_1}^{a_k} p_i(A)$$
\item The total potential payout, given by: $$\displaystyle \prod_{i=d_1}^{d_k} o_{i}^{j^*}(H) \times \displaystyle \prod_{i=h_1}^{h_k} o_{i}^{j^*}(D) \times \displaystyle \prod_{i=a_1}^{a_k} o_{i}^{j^*}(A)$$
\end{itemize}

\subsection{Model Formulation}
\label{5}
We first state the accumulator bet selection problem as a bi-objective mathematical program, in which three binary variables $x_{i}^j(H)$, $x_{i}^j(D)$, $x_{i}^j(A) \in \{0,1\}$ are associated to each match $m_i \in M$ and each bookmaker $b^j \in B$. These decision variables represent whether or not one should respectively bet on a home win, a draw, or an away win, in match $m_i$, using the services of bookmaker $b^j$. \\

A generic formulation of the problem is thus given by the following bi-objective mathematical program:
\begin{alignat}{4}
\max &\displaystyle \prod_{k\in \{H, D, A\}}  \prod_{i=1}^{n} \prod_{j=1}^{m}  (1 - (1-o_{i}^{j}(k))\cdot x_{i}^j(k))&  \\
\max & \displaystyle \prod_{k\in \{H, D, A\}}  \prod_{i=1}^{n} \prod_{j=1}^{m}  (1 - (1-p_{i}(k))\cdot x_{i}^j(k)) & \\
\text{s. t. }& \displaystyle \sum_{k\in \{H, D, A\}}  x_{i}^j(k) \leq 1 \quad \forall i \in \{1, \dots, n\}, \forall j \in \{1, \dots, m\} &\\
& \quad \displaystyle \sum_{j=1}^{m}  x_{i}^j(k) \leq 1 \quad \forall i \in \{1, \dots, n\}, \forall k \in \{H, D, A\}&\\
& x_{i}^j(k) \in \{0,1\} \quad \forall i \in \{1, \dots, n\}, \forall j \in \{1, \dots, m\}, \forall k \in \{H, D, A\}&
\end{alignat}

In this mathematical program, objective function $(1)$ seeks to maximize the total potential payout of the accumulator bet, when objective function $(2)$ seeks to maximize the estimated probability of this accumulator bet winning. These two polynomial functions have the same structure, in that if for a certain triplet $(i,j,k)$, decision variable $x_{i}^j(k)$ takes value $0$ in a solution, its corresponding term in the product would be equal to $1$ in each function, and the remaining terms would be considered, whereas $x_{i}^j(k)$ taking value $1$ would result in the odds of the corresponding individual bet being taken into account in function $(1)$ and its estimated probability being taken into account in function $(2)$. Note that these two functions can theoretically be reformulated linearly, using the \textit{usual linearization} technique described in \cite{LIBERTI}. However, this would entail introducing an additional binary variable and three constraints for each pair of initial decision variables, thus resulting in $2^{3nm -1}$ additional binary variables, and a proportional number of additional constraints, which would be impractical. Constraints $(3)$ state that, by definition, an accumulator bet cannot wager on two conflicting outcomes for a single match, and constraints $(4)$ imposes that a single bookmaker is chosen to place the wagers constituting an accumulator bet. Finally, constraints $(5)$ ensure that all variables are binary.

\subsection{Preprocessing}
The accumulator bet selection problem can be treated separately for each individual bookmaker, since all wagers constituting the accumulator bet have to be placed with the same bookmaker for the bet to hold. The problem can thus be divided, without loss of generality, into $m$ (the number of bookmakers) sub-problems, containing $m$ times less binary variables than the initial problem. In the perspective of solving the problem using a scalarizing function, finding the optimal solution of the initial problem would simply consist in comparing the $m$ optimal solution to the sub-problems, whereas in the perspective of enumerating the set of all efficient solutions, that would require testing for dominance between the Pareto sets of the sub-problems. \\

\noindent In addition to the above preprocessing step, which maintains it intact, the search space of the problem can be heuristically reduced by only considering variables $x_{i}^j(k), i \in \{1, \dots, n\}, j \in \{1, \dots, m\}, k \in \{H, D, A\}$ that correspond to single bets presenting Pareto-efficient pairs of odds and estimated probabilities  $(o_{i}^{j}(k), p_{i}(k))$, according to the two following dominance tests, which may however eliminate optimal or otherwise valuable combinations of bets:
\begin{itemize}
\item Intra-bookmaker dominance test: $\forall i \in \{1, \dots, n\}, \forall j \in \{1, \dots, m\}, \forall k \in \{H, D, A\}$, eliminate variable $x_{i}^j(k)$, if $\exists i' \in \{1, \dots, n\}$ and $\exists k' \in \{H, D, A\}: o_{i}^{j}(k) \leq o_{i'}^{j}(k')$ and $p_{i}(k) \leq p_{i'}(k')$, with at least one of these two inequalities being strict. In other words, an individual bet is eliminated from consideration, if there exists another individual bet with a higher probability, for which the same bookmaker offers higher odds. 
\item Inter-bookmaker dominance test: $\forall i \in \{1, \dots, n\}, \forall j \in \{1, \dots, m\}, \forall k \in \{H, D, A\}$, eliminate variable $x_{i}^j(k)$, if $\exists i' \in \{1, \dots, n\}$, $\exists j' \in \{1, \dots, m\}$ and $\exists k' \in \{H, D, A\}: o_{i}^{j}(k) \leq o_{i'}^{j'}(k')$ and $p_{i}(k) \leq p_{i'}(k')$, with at least one of these two inequalities being strict. In other words, an individual bet is eliminated from consideration, if there exists another individual bet with a higher probability, for which another bookmaker offers higher odds. 
\end{itemize}

The assumption underlying the intra-bookmaker dominance test is virtually made in any single bet selection model (which all consider a single bookmaker as well). It corresponds to the idea that selecting a non-Pareto-efficient individual bet is not rational, when other bets with better probabilities and odds are available. The inter-bookmaker dominance test is a generalization of this idea, when multiple bookmakers are considered. \\

However, it can be perfectly rational to select a dominated individual bet (according to either dominance tests), if the individual bets that dominate it are also part of the accumulator bet, as long as the resulting accumulator bet is Pareto-efficient. The experiment we present in Section \ref{hypo} will allow us to empirically evaluate the influence of these simplifying assumptions on the value of the generated accumulator bets.

\begin{figure}
\begin{center}
\includegraphics[width=0.75\linewidth]{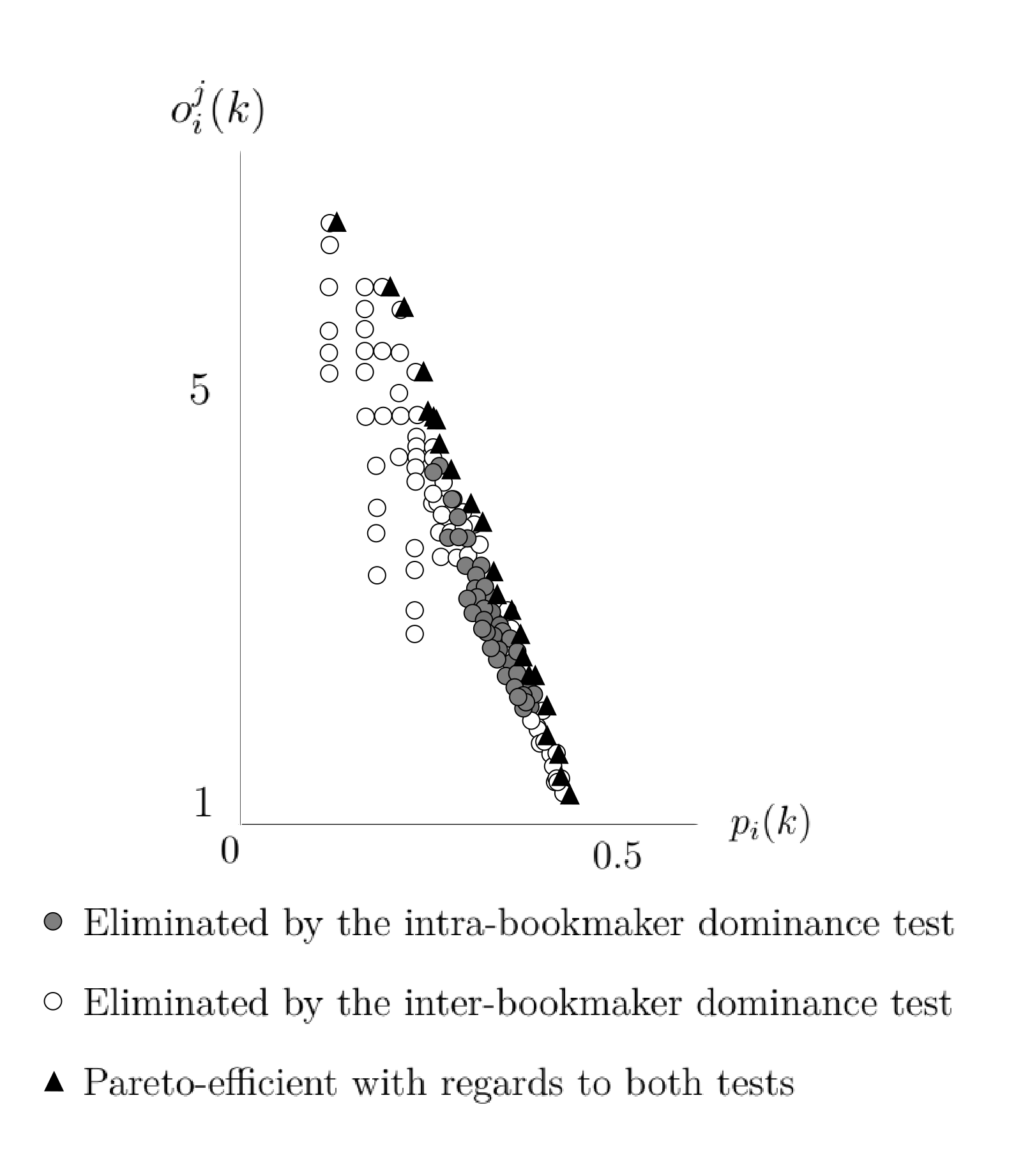}
\caption{\label{dominance} {Individual bets (circles) eliminated by intra and inter-bookmaker dominance tests. The remaining bets (triangles) are Pareto-efficient for both tests}} 
\end{center}
\end{figure}

We have estimated that over the 2015-2016 season (38 matches), when considering four leagues (\textit{LaLiga}, \textit{Premier League}, \textit{Bundesliga}, and \textit{Serie A}), five online bookmakers (namely \textit{Bet365}, \textit{Betway}, \textit{Gamebookers}, \textit{Interwetten}, and \textit{Ladbrokes}), the Intra-bookmaker dominance test reduces the number of variables (i.e. the number of individual wagers to consider) by an average of $64\%$, when the the Inter-bookmaker dominance test reduces this number by an average of $19\%$. Figure \ref{dominance} illustrates how these two dominance-tests can impact the set of possible individual bets for the first match day of the 2015-2016 season, when considering four leagues and five bookmakers.

\subsection{Scalarization}
Although it may be interesting to "solve" (i.e. identify and describe the Pareto set of) the accumulator bet selection problem in its bi-objective form, particularly in the perspective of implementing diversification strategies, one can more simply seek to maximize the odds of the selected accumulator, for a minimum probability $p_{min}$ (set at $25\%$ in this work), Therefore, in the remainder of this paper, we will consider the maximization of function $(1)$, under constraints $(3)$, $(4)$, and $(5)$, and the following additional constraint: 
\begin{alignat}{3} 
\max &\displaystyle \prod_{k\in \{H, D, A\}}  \prod_{i=1}^{n} \prod_{j=1}^{m}  (1 - (1-p_{i}^{j}(k))\cdot x_{i}^j(k))& \geq  p_{min}
\end{alignat}

\subsection{Stochastic Diffusion Search Algorithm}

In this Section we propose a Stochastic Diffusion Search (SDS) \cite{SDS1} algorithm to solve the problem defined by inequations $(3)$, $(4)$, $(5)$, and $(6)$, for each match day, and separately for each bookmaker. This multi-agent search approach is based on direct, one-to-one communication between agents and has been successfully applied to global optimization problems \cite{SDS15}. SDS has been shown to possess good properties in terms of convergence to global optima, robustness and scalability with problem sizes \cite{SDS2}, as well as a linear time-complexity \cite{SDS3}.\\

\begin{algorithm}[t]
Initialize agents\\
\textbf{repeat} \\
/* Test phase */\\
\For {all agents  $A_i$}
{Randomly choose an agent, $A_j$, from the population\\
\If{$Odds(A_i) \leq Odds(A_j)$ and $Prob(A_i) \leq Prob(A_j)$, \\
with at least one of these inequalities being strict}
{$Status(A_i)\leftarrow inefficient$

\textbf{else} \If{$Exp(A_i) < Exp(A_j)$}
{$Status(A_i)\leftarrow inactive$\\
\textbf{else}
$Status(A_i)\leftarrow active$

}
}
  }
  
  /* Diffusion phase */\\
\For {all agents  $A_i$}
{\If{$Status(A_i)=inefficient$}
{Reinitialize $A_i$\\
\textbf{else} \If{$Status(A_i)=inactive$}
{Randomly choose an agent $A_j$ from the population\\
\If{$Status(A_j)=active$}
{Set $A_i$ in the neighborhood of $A_j$\\
\textbf{else} Reinitialize $A_i$}

}
}
  }

\textbf{until} $\exists A_i: Exp(A_i) \geq min_{exp}$ or $Clock \geq max_{time}$  \\ 

 \caption{\label{SDS} Stochastic Diffusion Search Algorithm}
\end{algorithm}

After an initialization step in which each agent is given a candidate solution to the problem, an iterative search procedure is performed in two phases: 1. a test phase in which each agent compares the quality of its solution with that of another, randomly chosen, agent and 2. a diffusion phase in which agents share information about their candidate solutions on a one-to-one communication basis. This approach is formalized in algorithm \ref{SDS}.\\

Each agent $A_i$ is initialized with a potential accumulator bet, chosen by solving a relaxed, and simplified version of the problem that is constructed by setting a lower bound on function $(1)$ or function $(2)$, respectively corresponding to minimum acceptable odds or probability of a profit, and relaxing constraints $(5)$. We then solve the continuous mathematical program thus modified, for each bookmaker. For instance, a candidate solution can be generated by solving the following single-objective continuous mathematical program:
\begin{alignat*}{4}
\max &\displaystyle \prod_{k\in \{H, D, A\}}  \prod_{i=1}^{n} \prod_{j=1}^{m}  (1 - (1-o_{i}^{j}(k))\cdot x_{i}^j(k))&  \\
\text{s. t. }& \displaystyle \prod_{k\in \{H, D, A\}}  \prod_{i=1}^{n} \prod_{j=1}^{m}  (1 - (1-p_{i}(k))\cdot x_{i}^j(k)) \geq 0.25 & \\
& 0\leq x_{i}^j(k) \leq 1 \quad \forall i \in \{1, \dots, n\}, \forall j \in \{1, \dots, m\}, \forall k \in \{H, D, A\}&
\end{alignat*}
The resulting fractional solutions are then rounded to the nearest binary integer and violations of constraints $(3)$, i.e. betting on conflicting outcomes for the same match, are broken, by picking the outcome that presents the highest probability. We respectively denote $Odds(A_i)$, $Prob(A_i)$, the values of functions $(1)$, and $(2)$ for the solution carried by an agent $A_i$, and $Exp(A_i)=  Odds(A_i) \times Prob(A_i)$, its expected value. \\ 

During, the test phase, each agent $A_i$ compares the quality of its solution with that of a randomly chosen other agent $A_j$, and acquires one of three statuses, \textit{inefficient}, \textit{inactive}, or \textit{active}, corresponding to the three depending on the outcome of this comparison:
\begin{itemize}
\item The solution carried by agent $A_i$ is dominated by that of agent $A_j$, on both functions $(1)$ and $(2)$, that is to say that the accumulator bet assigned to agent $A_j$ presents both better total odds and a better probability of a win than those of the accumulator bet assigned to agent $A_i$, with at least one strict inequality. In this case, the status of agent $A_i$ is set to inefficient. 
\item The solution carried by agent $A_i$ is not dominated by that of agent $A_j$, with regards to function $(1)$ and $(2)$, but it presents a lower value on function $(1)\times(2)$, in other words the accumulator bet assigned to agent $A_j$ presents a better expected value than the accumulator bet assigned to agent $A_i$. In this case, the status of agent $A_i$ is set to inactive.
\item The solution carried by agent $A_i$ is not dominated by that of agent $A_j$, with regards to function $(1)$ and $(2)$, and presents a higher value on function $(1)\times(2)$. In this, case the status of agent $A_i$ is set to active.
\end{itemize}

In the diffusion phase, if an agent $A_i$ is inefficient it is reinitialized. If agent $A_i$ is inactive, a random agent $A_j$ is chosen. If $A_j$ is active, it communicates its hypothesis to $A_i$, by setting $A_i$ in the neighborhood of $A_j$. In this work, this operation is performed by replacing an individual bet in the accumulator carried by agent $A_j$, by a previously unselected, Pareto-efficient individual bet. However, if $A_j$ is inactive or inefficient, $A_i$ is reinitialized. Reinitialization is performed by selecting a random subset of three Pareto-efficient individual bets. The stopping condition of our algorithm is that an accumulator bet of a minimum preset expected value $min_{exp}$, given by function $(6)$ is encountered. In any given match day, if no solution satisfying this minimum requirement is generated after a duration $max_{time}$, no bet is placed on that match day.\\

{As pointed out in \cite{meta}, the efficiency of meta-heuristic algorithms depends on their ability to generate new solutions that can usually be more likely to improve on existing solutions, as well as avoid getting stuck in local optima. This often requires balancing two important aspects: intensification and diversification.
Similarly to the process introduced in \cite{SDS15} during the diffusion phase, our algorithm applies an intensification process around an active agent, to improve the efficient solution it carries, while a diversification process is used for inactive or inefficient agents to explore new solutions. }

\section{Application and experimental results}
\label{6}
\subsection{Data}
As stated in the introduction of this paper, we have considered a historical dataset, from the repository \url{http://www.football-data.co.uk}, and containing the results of each match day of the $2015-2016$ season, for four leagues (\textit{LaLiga}, \textit{Premier League}, \textit{Serie A} and \textit{Bundesliga}, the latter only being considered for possible picks up to match day 34), and the odds offered by five online bookmakers (namely \textit{Bet365}, \textit{Betway}, \textit{Gamebookers}, \textit{Interwetten}, and \textit{Ladbrokes}). The scope of our work being the combinatorial difficulty of the problem, we have used the popular Betegy service \url{http://www.betegy.com} for the assessments of the probabilities of these matches.
\subsection{Bankroll management}

We have used a conservative Kelly strategy \cite{KELLY} for money management over a season. In this strategy, the fraction of the current payroll to wager, on any optimal accumulator bet is $f= p - \frac{1-p}{o}$, where $o$ is the total potential payout of the accumulator bet and $p$ its probability of winning. However, contrary to the conventional Kelly criterion, we do not update the value of the total payroll after a gain, in order to limit potential losses.

\subsection{Tested hypotheses}
\label{hypo}
This experiment mainly intended to compare the performance of accumulator bets, under our model with individual betting strategies. We have implemented Rue and Salvesen's variance-adjusted approach \cite{9, 13} to generate recommendations regarding the latter type of bets. This approach minimizes the difference between the expected profit and its variance for each individual bet. If one invests a sum $c$ on a given individual bet with probability $p$ and odds $o$, this difference is given by $p\cdot o\cdot c - p(1 - p)(oc)^2$, and is minimum for  $c=\frac{1}{2 \cdot o\cdot(1-p)}$. This amount is thus wagered on each individual Pareto-efficient bet, and the cumulative profit, over a season, using this individual betting strategy was compared to that of our accumulator bet selection approach, combined with the conservative Kelly criterion we have adopted. The remaining two combinations of bet selection and bankroll management methods (i.e. accumulators with the variance-adjusted approach and single bets with the Kelly criterion) have been additionally considered for the sake of completeness, although they are relatively inefficient, as shown in Section \ref{res}. \\ 

Moreover, we have evaluated the opportunity losses due to simplifying our model using the intra-bookmaker dominance test. Thus each of our experiments was performed using both the initial model defined by inequations $(3)$ to $(6)$, and the simplified version in which non Pareto-optimal individual bets (according to the intra and inter-bookmaker dominance tests) have been eliminated.

{\noindent In this experiment, a minimum expected value $min_{exp}$ equal to $2$ has been empirically found to generate the best overall returns over a season. Lower values resulted in accumulator bets that offer no added value in terms of returns, compared to single bets, despite losing more frequently, while values higher than have been found to either generate accumulators with a large number of bets that are overly risky and unprofitable, or fail to generate any solution. Additionally, the maximum time limit $max_{time}$ has been arbitrarily set at $600$ seconds.}

\subsection{Results}
\label{res}
\begin{table}[!ht]
\hspace*{-.1\linewidth}

     \begin{center}
     \hspace*{-.1\linewidth}
     \begin{tabular}{|c|c|c|c|c|c|}

         \hline
     \mbox{\textbf{\small{Model}}} & \mbox{\textbf{\small{Preprocessing}}} & \mbox{\textbf{\small{Average odds}}} &\mbox{\textbf{\small{Average }}} &\mbox{\textbf{\small{Average stakes}}} & \mbox{\textbf{\small{Total Gains}}} \\
          &  & &\mbox{\textbf{\small{probability}}}&\mbox{\textbf{\small{per matchday}}} & \mbox{\textbf{\small{}}} \\
      \hline

         \mbox{Single betting}  & \mbox{N/A} & 2.87 &36\%  & 27.3\%      &30.1\%  \\
         \mbox{(variance adjusted)}  & \mbox{N/A} &  &  &      &  \\
               \cline{1-6}
     \mbox{Accumulator betting}        & \mbox{None} &83.1 &4.7\%  &   3.02\%     &
          
          37.6\%  \\
\mbox{(conservative Kelly)} & \mbox{Intra-bookmaker} & 83.1 & 4.7\% &   3.02\%    & 37.6\% \\
  
      & \mbox{Inter-bookmaker} & 90.2 & 4.2\% & 4.1\% & 12.9\%\\
            \cline{1-3}
             \hline   
             
       \end{tabular}
        \caption{\label{results} Gains over the $2015-2016$ season using the four models}
          \end{center}
\end{table}
The results for four betting strategies over the $38$ matches of the $2015-2016$ are given in Table \ref{results}. Gains over the full simulation are represented as a percentage of the bettor's total bankroll. The first observation is that the initial binary optimization model and its simplified version using the intra-bookmaker dominance test produce exactly the same recommendations and thus the same results. We can see that the influence of the simplifying hypothesis we have made with the intra-bookmaker dominance test is neglectable. Eliminating individual bets that are dominated according to the inter-bookmaker dominance test may however be limiting, as we can observe a decrease in value compared to the initial accumulator betting model. \\

\begin{figure}[!h]
\begin{center}
\centering

\hspace*{-.15\linewidth}
\includegraphics[width=1.3\linewidth]{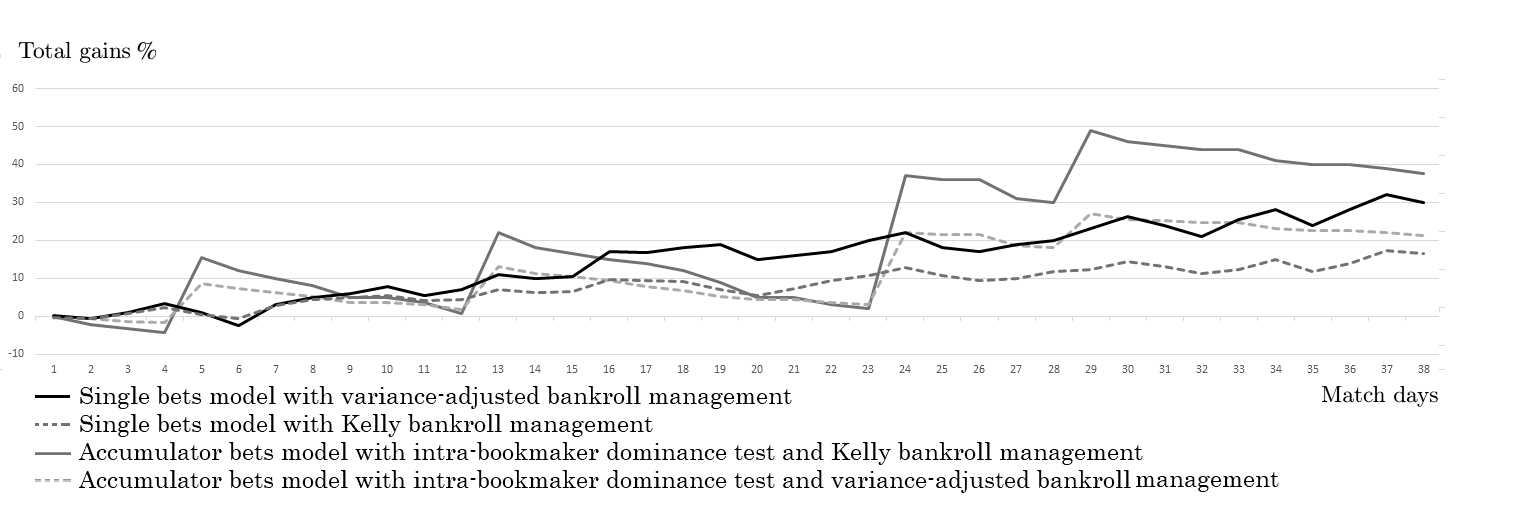}
\caption{\label{results2} Cumulative gains per match day, over the $2015-2016$ season}
\end{center}
\end{figure}

Table \ref{results} additionally presents the average odds and probabilities of the single and accumulator betting models considered in this experiment, which highlights their markedly different nature. Generated accumulator bets with the intra-bookmaker dominance test present average odds of $83.1$ and average probability of $4.7\%$, when single bets have average odds of $2.87$ and average probability of $36\%$. This translates into average stakes per match day of markedly different orders for the two approaches ($27.3\%$ of the bettors bankroll for the single betting model and $3.02\%$ of this capital for the accumulator betting model). 

\begin{figure}[!h]
\begin{center}
\centering

\hspace*{-.15\linewidth}
\includegraphics[width=1.3\linewidth]{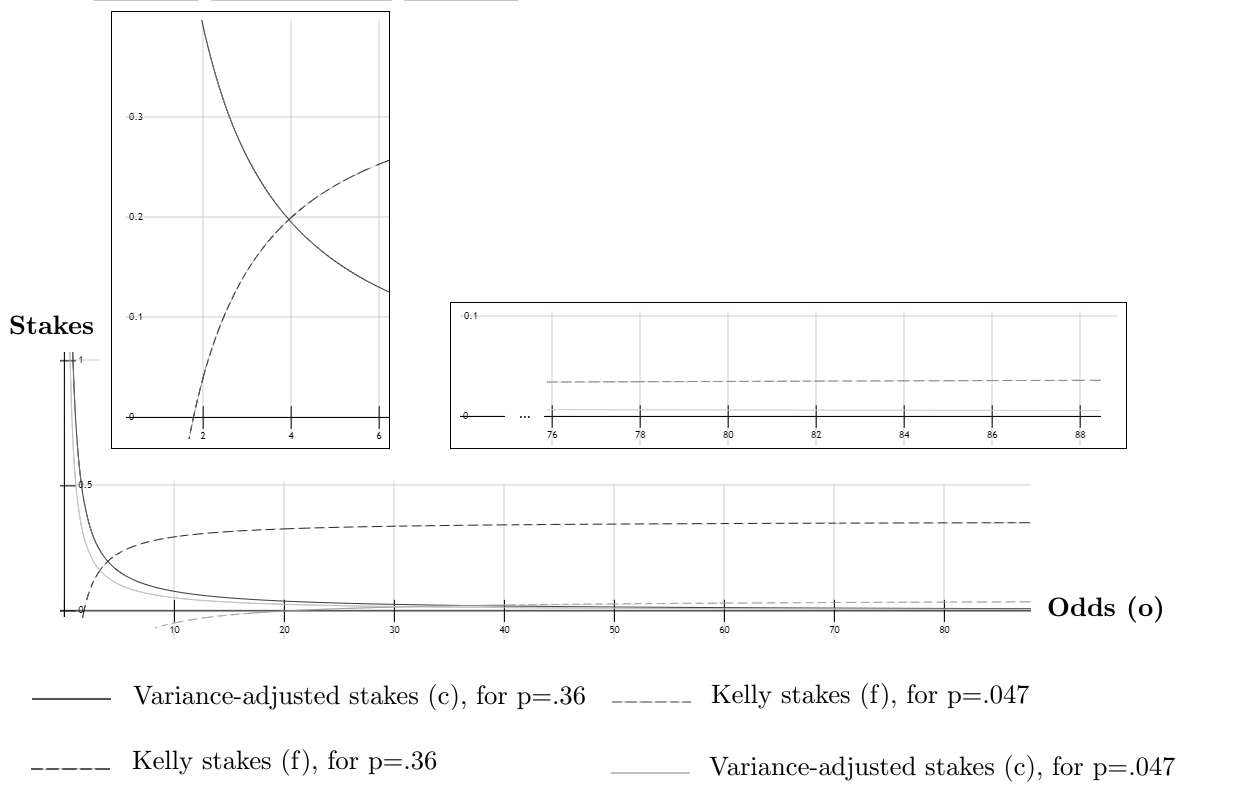}
\caption{\label{functions} Kelly and variance-adjusted stakes for the average probabilities of single and accumulator bets, as functions of the odds} 
\end{center}
\end{figure}
These differences result in different behaviors for the two types of models. Given their much lower stakes, for a single bets selection model to be profitable, relatively high stakes need to be wagered every match day and the model would seek to maximize the number of winning bets among these wagers, when an successful accumulator betting strategy would seek to limit the extent of its frequent losses to ensure the survival of the bettor for the rare match days in which the model is right. The differences of orders of magnitude between the average odds and probabilities of the considered accumulator and those of single bets also explain why the former is more efficient when combined with the Kelly criterion, when the later seems to be more adapted to a variance-adjusted bankroll management strategy, as can be seen in Figure \ref{results2}, detailing the variation of the bankroll of our bettor, over the $38$ match days constituting the season. Indeed, the Kelly criterion consistently bets lower stakes for values of probabilities and odds characteristic of single bets, when there is a reversal of this order for probabilities and odds characteristic of accumulator bets and the variance-adjusted function bets significantly lower stakes than the Kelly criterion. Figure \ref{functions} represents the respective stakes $f$ and $c$ of the Kelly and variance-adjusted bankroll management strategies as a function of the odds of a bet (single or accumulator), for the average estimated probability of the two types of bets in our experiment (respectively at $p=.36$ and $p=.047$). This figure additionally emphasizes the stakes given by the two functions for odds of values around the average odds of each type of bets (respectively at $o=2.87$ and $o=83.1$, for single bets and accumulators). Over a season, these differences result in the less efficient combination of approach being a smoothed version of the dominant one as can be observed in Figure \ref{results}. In the case of accumulator bets, the Kelly bankroll management strategy in which the bankroll is not updated after gains is already very conservative, with average stakes per match day at $3.02\%$. Further restricting these stakes through the use of the variance-adjusted function would unnecessarily limit potential gains. The same effect has been observed, in this experiment, for the Kelly strategy combined with single bets. However, since the stakes function here is applied to each individual bet, the relationship between the two gains curves is less systematic than in the case of accumulators. It is for instance, possible for the Kelly criterion to result in an overall gain in a given match day, when the variance-adjusted criterion results in an overall loss. \\

For any given bankroll management strategy, we can observe that the growth in net returns using single bets models is more regular than when using an accumulator bets models. The latter strategy typically loses money for several weeks in a row, before experiencing growth by spurts, corresponding to a low-probability/high-odds accumulator bet winning. Losses are however limited by an adequate bankroll management strategy, and in the most profitable version of this model, only four winning accumulator bets over the season were sufficient to ensure a $37.6\%$ total return. Comparing the two most profitable strategies in this experiment, and although we have considered a limited number of leagues and bookmakers, the accumulator betting model with preprocessing through the intra-bookmaker dominance test combined with a Kelly bankroll management strategy has outperformed the single bet model combined with a variance-adjusted money management strategy, over the $2015-2016$ season and can be profitable betting approach.
\section{Conclusion}
\label{7}
The relatively under-studied problem of accumulator bet selection presents a rather uncommon combinatorial structure, due to the multiplicative nature of its returns. This work proposed a model and solving approach for the treatment of this problem. Preliminary experiments showed promising results in terms of payouts relative to traditional single bet selection approaches. The model can be used to create commercial automated recommendation programs for bookmaker offices. However, the proposed betting strategy requires the decision-maker to be willing to accept losses over multiple periods of time before a gain, which may render the model inadequate for certain types of users. There is thus a a twofolds challenge in designing such a strategy. The decision of when not to not bet (or what limit to put on the sum wagered) is as important as that of placing a bet. In our approach, this aspect is taken into account at the micro level of a match day, by setting a minimum expected gain requirement, as a stopping condition for the stochastic diffusion search, and at the macro level of money management over a season, by following a conservative Kelly strategy. {Regarding the former aspect, the choice of the value of the variable representing the minimum expected gain is an aspect that certainly warrants further experimentation, with different datasets, given how crucial it is for this betting strategy.} Further refinements in the model are also required regarding the money management strategy over a season. Lastly, solving the problem in its original bi-criteria form and generating the Pareto front of non-dominated accumulator bets can constitute a sufficient recommendation for more expert bettors to make a choice from. It could thus prove fruitful to design algorithms for the treatment of the problem in this form. 

\bibliographystyle{apalike}
\bibliography{ref}

\end{document}